\documentclass[10pt,twocolumn,letterpaper]{article}

\usepackage[pagenumbers]{cvpr}              
\usepackage{comment}
\usepackage{multirow}
\usepackage{xcolor}

\usepackage[para]{footmisc}

\definecolor{cvprblue}{rgb}{0.21,0.49,0.74}
\usepackage[pagebackref,breaklinks,colorlinks,allcolors=cvprblue]{hyperref}

\newcommand{\methodname}{\textbf{GenMask}\xspace}

\def\paperID{5079} 
\def\confName{CVPR}
\def\confYear{2026}

\title{\vspace{-1.5em}\methodname: Adapting DiT for Segmentation via Direct Mask Generation\vspace{-0.5em}}

\author{
Yuhuan Yang$^{1,4}$\thanks{Work done while the author was an intern at Alibaba Group. Email: {yangyuhuan@sjtu.edu.cn}}\quad
Xianwei Zhuang$^{4}$\quad
Yuxuan Cai$^{4}$\thanks{Project Leader}\quad
Chaofan Ma$^{1}$\quad
Shuai Bai$^{4}$\quad
Jiangchao Yao$^{1,2}$\\
Ya Zhang$^{2,3}$\thanks{Corresponding authors. Email: {ya\_zhang@sjtu.edu.cn, wangyanfeng622@sjtu.edu.cn}}\quad
Junyang Lin$^{4}$\quad
Yanfeng Wang$^{2}$\footnotemark[\value{footnote}]\\[0.5em]
$^{1}$Cooperative Medianet Innovation Center, Shanghai Jiao Tong University\\
$^{2}$School of Artificial Intelligence, Shanghai Jiao Tong University\\
$^{3}$Institute of Artificial Intelligence for Medicine, Shanghai Jiao Tong University School of Medicine\\
$^{4}$Alibaba Group
}

\begin{document}
\maketitle
\begin{abstract}
    Recent approaches for segmentation have leveraged pretrained generative models as feature extractors, treating segmentation as a downstream adaptation task via indirect feature retrieval. 
    This implicit use suffers from a fundamental misalignment in representation.
    It also depends heavily on indirect feature extraction pipelines, which complicate the workflow and limit adaptation.
    In this paper, 
    we argue that instead of indirect adaptation, 
    segmentation tasks should be trained directly in a generative manner.
    We identify a key obstacle to this unified formulation: VAE latents of binary masks are sharply distributed, noise robust, and linearly separable, distinct from natural image latents.
    To bridge this gap, we introduce timesteps sampling strategy for binary masks that emphasizes extreme noise levels for segmentation and moderate noise for image generation, enabling harmonious joint training.
    We present \methodname, a DiT trains to generate black-and-white segmentation masks as well as colorful images in RGB space under the original generative objective. \methodname preserves the original DiT architecture while removing the need of feature extraction pipelines tailored for segmentation tasks. 
    Empirically, \methodname attains state-of-the-art performance on referring and reasoning segmentation benchmarks and ablations quantify the contribution of each component.
\end{abstract}
    
\section{Introduction}

Text-based segmentation is an important problem in computer vision.
It requires the model to predict a binary mask based on natural language descriptions of the image content.
With the recent emergence of large-scale self-supervised discriminative pretraining, it is increasingly treated as a downstream adaptation task rather than being learned from scratch.
Models such as CLIP~\cite{clip}, trained on web-scale uncurated data,
have demonstrated exceptional capability in capturing high-level visual semantics, 
thereby offering strong initialization for a variety of segmentation frameworks~\cite{ghiasi2022scaling,cris,risclip,xu2022groupvit,ma2023attrseg, ma2022fusioner,yang2024multimodal,zhang2023prototypical}.

Meanwhile, the rapid progress of text based image generation models, especially large scale pretrained latent diffusion models~\cite{stable_diffusion,stable_diffusion_3_5}, has sparked growing interest, 
and representations behind them are also widely explored for various of vision tasks including text-based segmentation~\cite{li2023your,tang2023emergent,odice_diffusion}.
Following the paradigm of using discriminative pretrained models, these works typically treat pretrained diffusion generative models as backbones.
Segmentation masks are obtained by first extracting hidden features during the denoising or diffusion-inversion process, and then feeding the extracted features into a trainable task-specific decoder~\cite{fundel2025distilldift,lee2024dmp,meng2024diffusionmodelactivationsevaluated,luo2023dhf,yang2023diffusion,stracke2025cleandift,ma2025freesegdiff}.

\begin{figure*}[!htp]
    \centering
    \includegraphics[width=\linewidth]{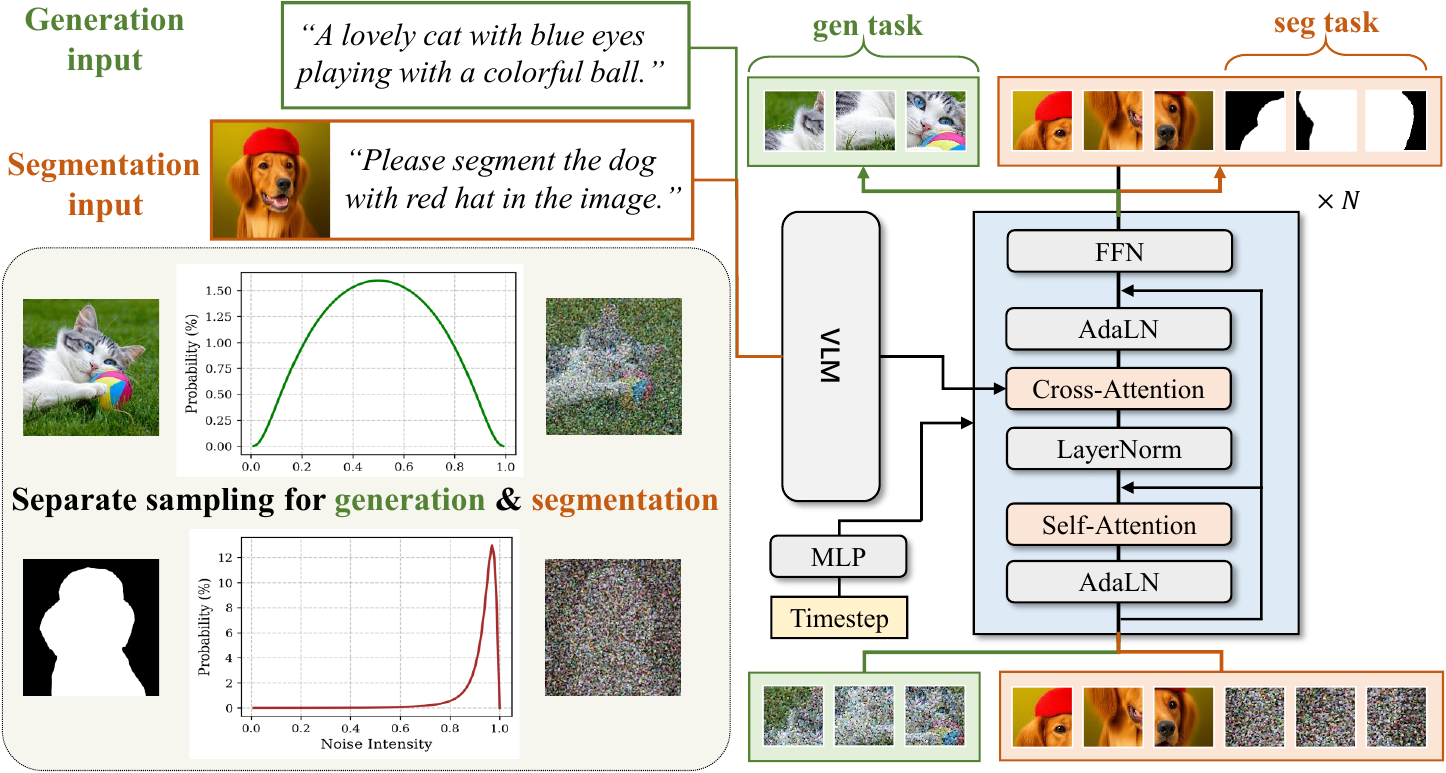}
    \caption{\textbf{Overall architecture of our model.}
    Here \textcolor{brown}{\textbf{brown}} and \textcolor{ForestGreen}{\textbf{green}} denote the {segmentation} and {generation} data flow respectively. 
    \textcolor{ForestGreen}{\textbf{Generation}} task follows standard diffusion training process, and its timesteps sampling strategy is similar to Stable Diffusion 3 \cite{Esser2024SD3}, emphasizing on intermediate denoising steps.
    For \textcolor{brown}{\textbf{segmentation}} task, we use an extreme long tailed sampling strategy.
    We also add VAE representation of the input image into DiT to supplement low-level information for segmentation.
    }
    \vspace{-1em}
    \label{fig:architecture}
\end{figure*}

Despite progress, these works still rely on an \textit{implicit use} of pretrained diffusion models, and therefore suffer from two key limitations.
\textbf{(1)} Diffusion models are pretrained to model the low-level distribution of VAE features, whereas segmentation requires compact, semantic-level label predictions.
This representational mismatch hampers effective downstream adaptation.
\textbf{(2)} Existing methods rely on carefully designed, indirect pipelines to extract features from diffusion models. Common approaches include diffusion inversion~\cite{ma2023diffusionseg} and activation aggregation~\cite{yang2023diffusion,meng2024diffusionmodelactivationsevaluated}. 
These intermediate operations also complicate the workflow and limit adaptation performance.

In this paper, we argue that instead of indirect adaptation, \textit{segmentation tasks should be trained directly in a generative manner}.
Our method, \methodname, realizes this idea by training a Diffusion Transformer (DiT) to directly generate black-and-white segmentation masks in RGB space under a generative training objective.
By doing so, we demonstrate three distinct merits. 
\textbf{(1)} Architecturally faithful to the original DiT. The segmentation process can be integrated into the original end-to-end DiT framework without structural changes or extra operations. 
\textbf{(2)} Maximally aligned with the generative training objective. The model continues to be trained under a generative objective, eliminating the optimization gap caused by implicit adaptation. 
\textbf{(3)} Seamless incorporation of generated data. Generation and segmentation can be trained jointly, allowing the use of generative data to improve segmentation performance.

Specifically, we cast text-to-image generation and text-based segmentation as a single conditional generation objective: the model learns to produce either an image or a segmentation mask with given condition.
While pursuing this unified formulation, we discover that \textit{a large gap exists} between the VAE representations of binary segmentation masks and those of natural RGB images.
VAE features for RGB images are smooth and easily perturbed by Gaussian noise, whereas features for binary masks are {sharply distributed, highly robust to noise, and largely linearly separable}.
This representational discrepancy makes it difficult for a single generative model to learn both distributions well simultaneously.
To address this, we introduce a specific \textit{timesteps sampling strategy} for segmentation masks: we sample extremely high noise levels more frequently, while for generation examples we emphasize moderate noise levels. 
This tailored sampling lets the model capture the two distinct feature distributions effectively.
We further optimize the inference pipeline to produce masks with a single model forward pass. 
As a result, we obtain a deterministic segmentation model trained under the same generative objective in pretraining.

Based on this solution, we build our model on a pretrained DiT and employ a vision-language model (VLM) to encode both visual and textual instructions for generation and segmentation. 
For segmentation, we also inject the input image's VAE latent as a low-level shortcut to provide the texture and color cues needed for accurate pixel-level prediction. 
Beyond achieving state-of-the-art results on referring and reasoning segmentation benchmarks, we also present comprehensive empirical studies that quantify the contribution of each key component in our architecture.

\vspace{-0.1em}
\section{Method}
\vspace{-0.1em}

We first provide the necessary preliminaries on the diffusion algorithm and model architecture in \cref{sec:overview}.
Then, \cref{sec:algorithm} presents our key contribution: sampling strategy that integrates binary segmentation mask into the conventional low-level visual sampling process of the diffusion model.
Finally we introduce the detailed implementation of our architecture in \cref{sec:architecture}, 
describing how we harmonize the discriminative segmentation task with the low-level visual generation dynamics in a unified learning paradigm.

\subsection{Overview}
\label{sec:overview}
\noindent\textbf{Preliminaries on Flow Matching.}
Flow Matching~\cite{lipman2022flow} is a generative modeling framework that learns a continuous path to transform simple noise (e.g., Gaussian) into complex data (e.g., natural images).
An effective version of Flow Matching is Rectified Flow~\cite{liu2022flow}. Instead of using complex paths, it trains the model with the simplest one: a straight line between data and noise.
Concretely, for each image $\mathbf{x}_0$ and a random Gaussian noise $\mathbf{\epsilon}\sim N(0,\mathbf{I})$, we define a linear path that connects $\mathbf{x}_0$ to $\mathbf{\epsilon}$ with a constant direction vector $\mathbf{v} = \mathbf{x}_0 - \mathbf{\epsilon}$ over the time interval $t\in[0,1]$ :
\begin{equation}
  \mathbf{x}_t = \mathbf{x}_0 - t\cdot \mathbf{v} = t\mathbf{\epsilon} + (1-t)\mathbf{x}_0.
\end{equation}
The goal of Rectified Flow is to train a neural network $v_\theta(\mathbf{x}_t,t)$ to predict the direction vector $\mathbf{v}$. And the loss is defined as:
\begin{equation}
  \mathcal{L}(\theta) = \mathbb{E}_{\mathbf{x}_0, \mathbf{\epsilon}, t} \left\| (\mathbf{x}_0 - \mathbf{\epsilon}) - v_\theta(\mathbf{x}_t,t)\right\|^2.
  \label{eq:mse-loss}
\end{equation}

\noindent\textbf{Architecture Overview.}
\methodname integrates both text-to-image generation and language-guided segmentation tasks in one framework without additional parameter. 
As illustrated in ~\cref{fig:architecture}, both tasks rely on the same diffusion training process.
The only variation between them is the timesteps sampling schedule: segmentation uses an aggressively long-tailed distribution to focus learning on the high-noise region.

\subsection{Timesteps Sampling for Segmentation Masks}
\label{sec:algorithm}

Natural images contain rich textures, diverse colors, and fine-grained details, whereas binary segmentation masks contain only sparse foreground–background patterns and possess extremely low visual complexity.
Due to this discrepancy, the latent space of masks occupies a narrow, highly biased region, making it difficult for generative models trained on natural-image distributions to model mask statistics reliably.
Such distributional mismatch also underlies the limitations of previous diffusion model based segmentation approaches.
In this section, we first highlight this inherent bias and then present our timesteps sampling strategy for segmentation mask to address this challenge.
Specifically, in \cref{sec:seg_distribution}, we examine in detail how the distribution of segmentation masks differs from that of natural images.
Then, in \cref{sec:gen_sample}, we review the widely adopted timesteps sampling strategy for image generation.
Finally we introduce our sampling strategy for segmentation task in \cref{sec:seg_sample}.
By \textit{separating general image denoising and mask denoising into different timesteps with different noise intensity}, the model can learn two tasks simultaneously with a unified architecture and training objective.

\begin{figure}[htp]
  \centering
  \includegraphics[width=\linewidth]{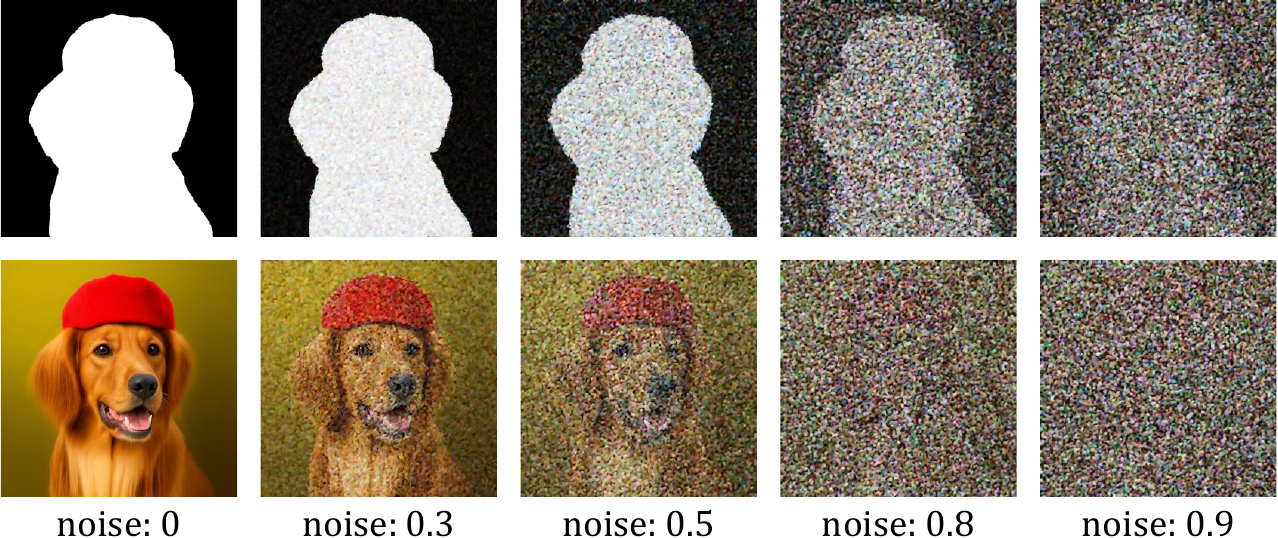}
  \caption{\textbf{The process of adding noise to natural image and binary mask.}
  Compared with natural image, binary mask is much more robust to noise.
  }
  \label{fig:denoise1}
\end{figure}

\subsubsection{Latent Distribution for Binary Masks}
\label{sec:seg_distribution}

We starts from a demo example by visualizing the process of adding noise to a natural image and a binary mask in ~\cref{fig:denoise1}.
Interestingly, we find that \textit{binary masks are much more robust to noise} than natural images.
For a natural image, introducing an extremely high level of noise completely obliterates its content, making the result almost indistinguishable from random noise. 
In contrast, when the same noise level is applied to a binary segmentation mask, the global position and shape of the segmented region remain largely intact, even the boundaries remain clearly recognizable.
These observations suggest that the latent representations of binary masks may be fundamentally different from those of natural images.

\begin{figure}[thp]
  \vspace{-0.2em}
  \centering
  \begin{subfigure}[b]{0.3\linewidth}
    \centering
    \includegraphics[width=\linewidth]{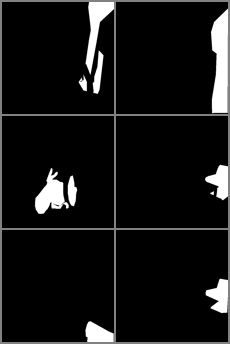}
    \caption{}
  \end{subfigure}\hfill
  \begin{subfigure}[b]{0.3\linewidth}
    \centering
    \includegraphics[width=\linewidth]{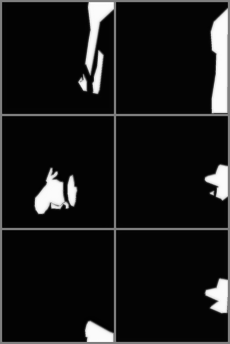}
    \caption{}
  \end{subfigure}\hfill
  \begin{subfigure}[b]{0.3\linewidth}
    \centering
    \includegraphics[width=\linewidth]{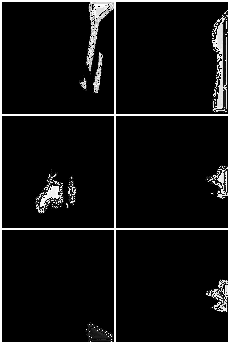}
    \caption{}
  \end{subfigure}
  \vspace{-0.3em}
  \caption{\textbf{VAE features for binary segmentation masks are linearly separable.} 
  \textbf{(a)} The input segmentation masks. 
  \textbf{(b)} PCA label of its VAE representation. 
  \textbf{(c)} The difference between the input masks and the PCA label after histogram normalization.}
  \label{fig:pca}
  \vspace{-1em}
\end{figure}

To further understand this phenomenon, we analyze a toy example and uncover a simple yet often overlooked fact: \textit{the VAE representation of binary segmentation masks is effectively linearly separable}.
We randomly sample $N$ segmentation masks from our dataset, and encode them into VAE representation with shape $
\mathbf{X}\in\mathbb{R}^{N\times hw\times d}$. Here $d=16$ is the VAE latent dimension.
We see $\mathbf{X}$ as $N\times hw$ data, and perform PCA decomposition into only ONE principal component, and gets $\mathbf{Y}=\mathbf{X}\mathbf{W}\in\mathbb{R}^{N\times hw}$. The whole process is shown as:
\begin{equation}
  \begin{array}{cc}
    \text{Binary Mask} \\ N \times HW
  \end{array}
  \to
  \begin{array}{cc}
    \text{VAE Feature} \\ N \times hw \times d
  \end{array}
  \to
  \begin{array}{cc}
    \text{PCA Label} \\ N \times hw
  \end{array}
\end{equation}
We take $N=100$ and shows 6 of the input mask and output the visualization for both segmentation mask and PCA label in ~\cref{fig:pca}. We find the PCA label is extremely similar with input mask, which means, the VAE representation space is linear separable with $\mathbf{W}$.

\begin{figure}[thp]
  \centering
    \includegraphics[width=0.9\linewidth]{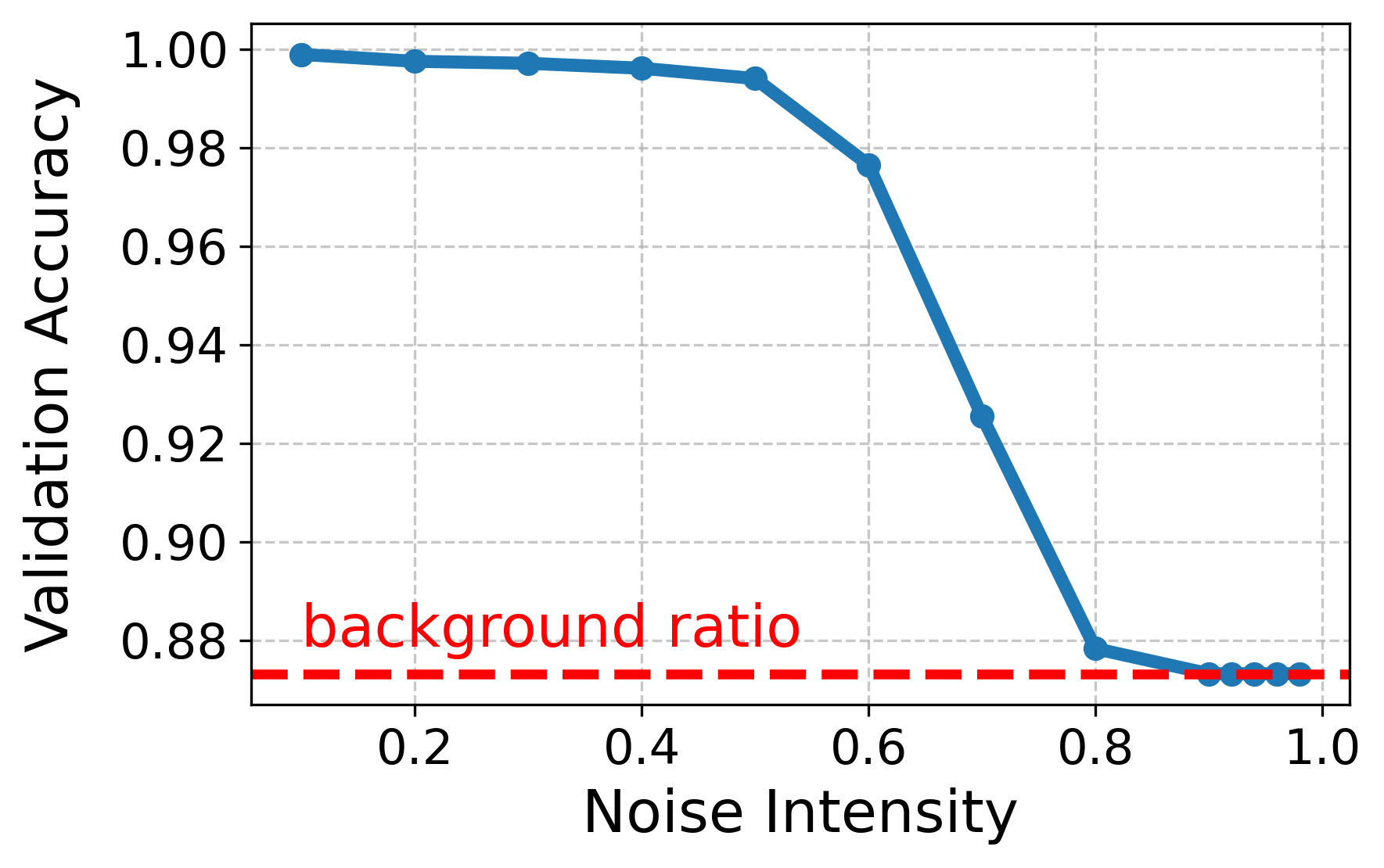}
  \vspace{-0.5em}
    \caption{\textbf{SVM validation accuracy on VAE embeddings of binary masks under different noise levels.}
    The embeddings remain linearly separable under low noise, while only high-intensity perturbations substantially degrade separability.
    }
    \label{fig:svm_results}
    \vspace{-1em}
\end{figure}

Finally, we gradually add noise to the VAE representation of the input mask and use least squares classification for label regression. 
The validation accuracy is shown in~\cref{fig:svm_results}. 
The results reveal that only at high noise intensity does the linear separability collapse, providing meaningful information for segmentation.

\subsubsection{Time Shift for Generation}
\label{sec:gen_sample}
The non-uniform importance of denoising steps has been studied for image generation task.
For generation task, early timesteps (dominated by noise) and late timesteps (concerned only with fine details) provide limited useful learning signals. 
Stable Diffusion 3 (SD3)~\cite{Esser2024SD3} proposes a resolution-dependent timesteps sampling strategy.
Following SD3, we use the logit-normal sampling strategy to emphasize intermediate noise levels during training. The probability density function of timestep $t$ is given by:
\begin{equation}
  \pi(t) = \frac{1}{\sqrt{2\pi} t(1-t)}\exp\left(-\frac{1}{2}\left[\log\left(\frac{t}{1-t}\right)\right]^2\right).
  \label{eq:gen-sample}
\end{equation}
In practice, we sample the random variable $u$ from a normal distribution $u\sim\mathcal{N}(0, 1)$, and then transform it to timestep $t$ using the inverse of the cumulative distribution function:
\begin{equation}
  t = \frac{1}{1+e^{-u}}.
\end{equation}

\begin{figure}[thp]
  \vspace{-1em}
  \begin{subfigure}[b]{\linewidth}
    \centering
    \includegraphics[width=0.9\linewidth]{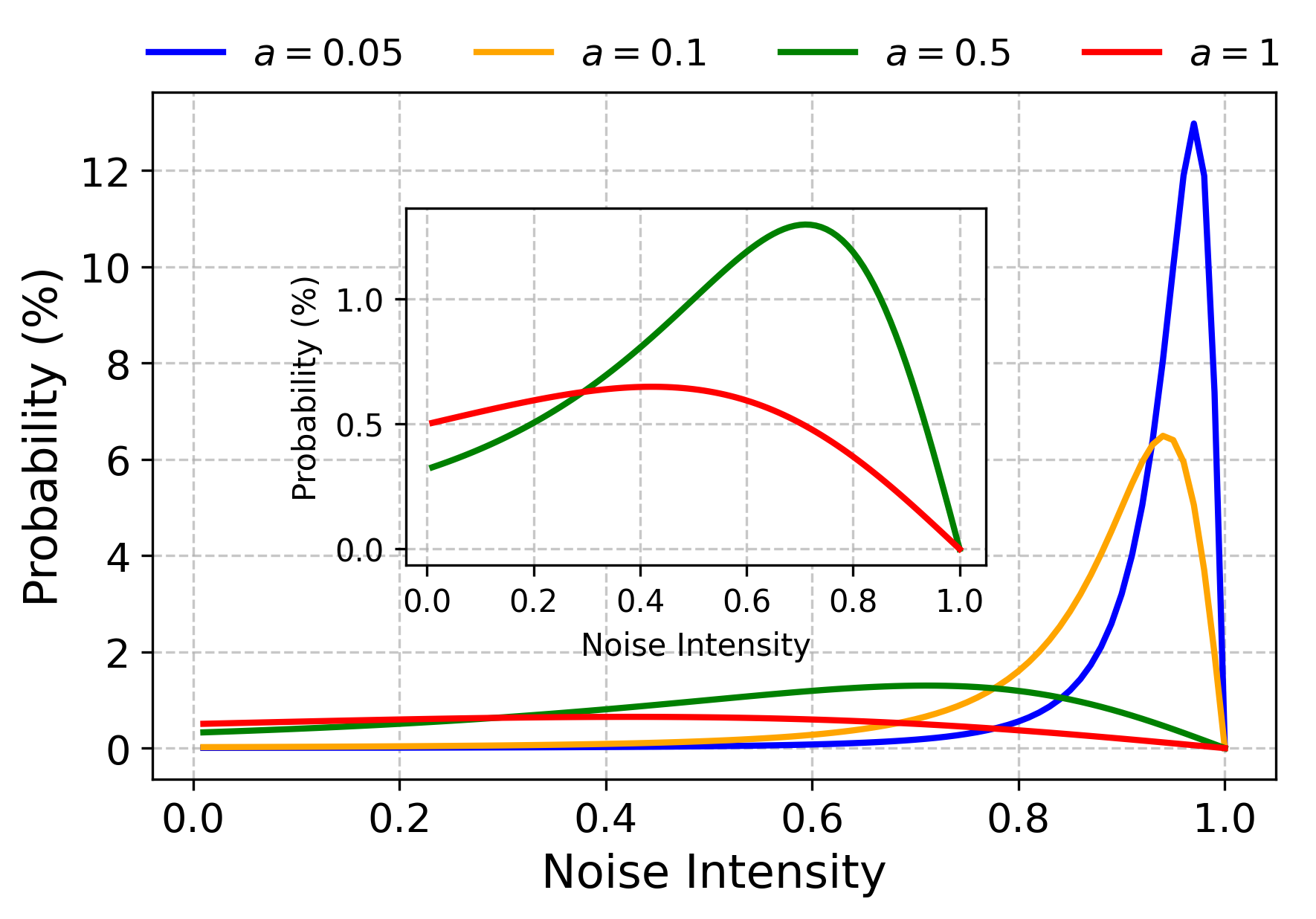}
  \end{subfigure}
  \begin{subfigure}[b]{\linewidth}
    \centering
    \includegraphics[width=0.9\linewidth]{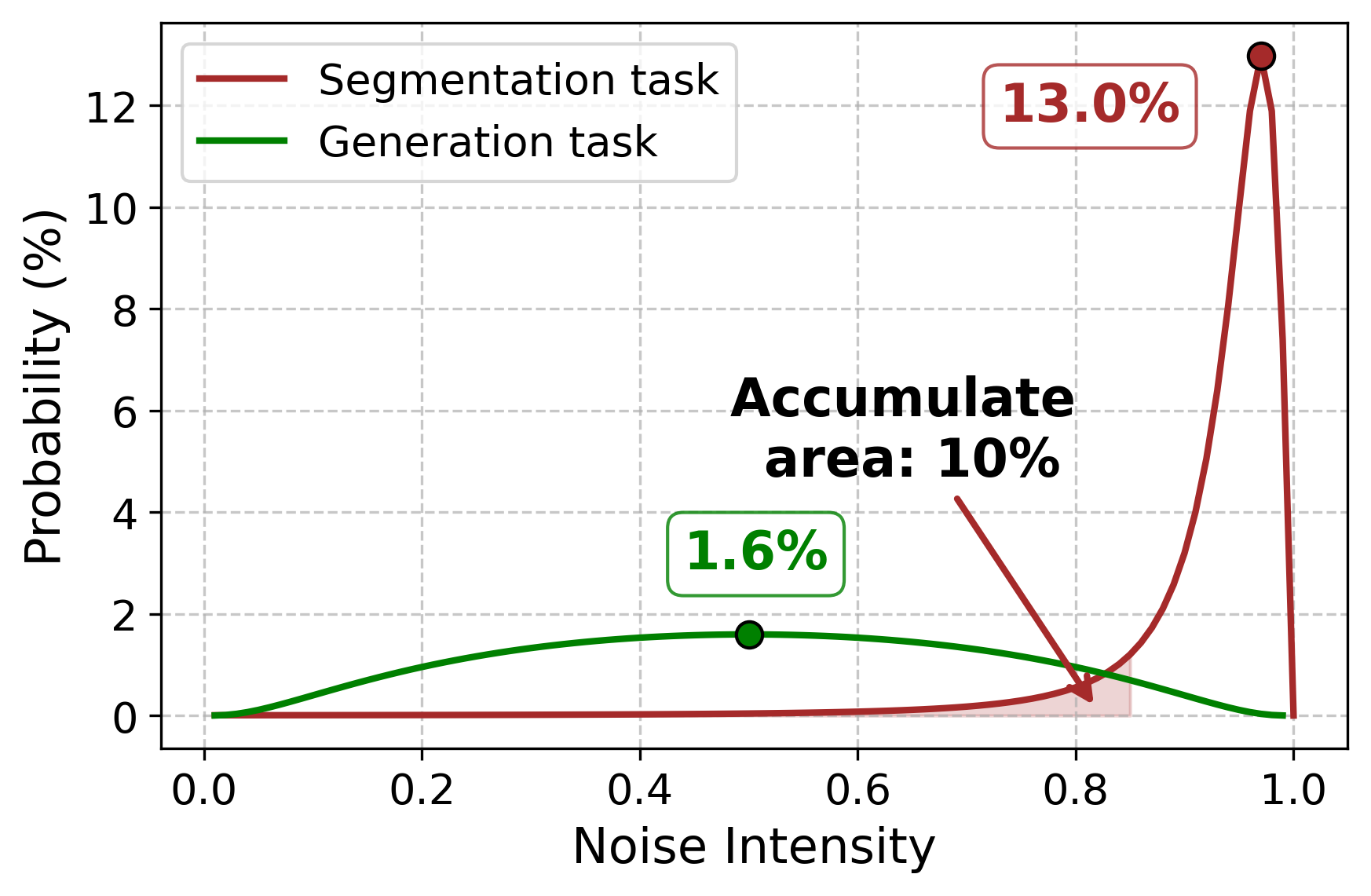}
  \end{subfigure}
  \vspace{-1.5em}
  \caption{\textbf{(Up) Importance resampling function for segmentation task with different hyperparameter $a$.} More extreme $a$ value means the distribution is more concentrated in high noise intensity regime. \textbf{(Down) Different sampling strategies for segmentation and generation task separately.} Generation task uses a relatively uniform sampling strategy, only emphasize the intermediate denoising steps. While segmentation task uses a extreme long tailed sampling strategy, with peak value 8× higher.}
  \label{fig:sampling2_3}
  \vspace{-1.5em}
\end{figure}

\subsubsection{Time Shift for Segmentation}
\label{sec:seg_sample}
Inspired by the timesteps sampling strategy for generation task, we propose that segmentation task also needs a tailored sampling strategy during training to ensure effectiveness.
This sampling strategy should be long tailed, and concentrate in the high noise intensity regime.
Here we construct a probability density function $p(t)$ with extreme long tail in early timesteps: 
\begin{equation}
  p(t) = \frac{2a^2t}{(t^2+a^2)^2}.
  \label{eq:seg-sample}
\end{equation}
And in practice, we first sample the uniform distributed random variable $u\sim\mathcal{U}(0, 1)$, and then transform it via:
\begin{equation}
  t = \sqrt{\frac{u}{1-u}}\cdot a,
\end{equation}
where $a$ is a hyperparameter about time shift. ~\cref{fig:sampling2_3} (Up) shows the distribution curve with different $a$. 
Smaller $a$ means the distribution is more concentrated in high noise intensity regime.
We draw the two sampling strategies from \cref{eq:gen-sample} and \cref{eq:seg-sample} together in ~\cref{fig:sampling2_3} (Down), and observe that the noise distributions for the two tasks are completely different.
The generation task adopts a near-uniform sampling strategy, where the probability mass around the intermediate noise region is only slightly elevated, resulting in a modest peak of 1.6\%. This mild adjustment essentially equivalent to adding a small weight to mid-range timesteps during training~\cite{Esser2024SD3}.
By contrast, the segmentation task relies on an extremely long-tailed distribution with a pronounced peak of 13\%, over 8× higher than that of the generation task. The cumulative probability below t = 0.85 is merely 10\%, meaning that 90\% of training samples are intentionally concentrated in the high-noise region.

\subsubsection{One-step Inference for Segmentation}
\label{sec:inference}
Since the segmentation task is trained predominantly on high–noise-intensity timesteps, low-noise regions provide only limited discriminative information for mask prediction.
This property allows us to bypass the multi-step progressive denoising steps, which are typically required in diffusion inference.
During inference, we fix the sampling timestep $t=1$, the segmentation mask is generated with only one model forward pass:
\begin{equation}
  x_{\text{mask}} = \epsilon + v(\epsilon, 1).
\end{equation}
Finally, we decode the latent representation $x_{\text{mask}}$ with VAE decoder to get the final mask.

Remarkably, in usage pattern, this one-step decoding process aligns perfectly with those conventional, carefully designed segmentation decoders, yet it requires no changes in original diffusion network architecture or additional training parameters. 
This reveals an appealing property of our model: despite with purely generative training objective, it naturally yields deterministic and accurate segmentation, aligning seamlessly with the demands of real-world deployment.

\subsection{Model Architecture and Training Objectives}
\label{sec:architecture}

\methodname is built upon the pretrained WAN-2.1 DiT~\cite{wan2025} architecture. %
\cref{fig:architecture} illustrates the overall architecture of our proposed model.

\noindent\textbf{DiT Decoder.}
WAN-2.1 is a cross-attention based DiT. 
It accepts the noisy image as input, then uses cross-attention mechanism to integrate the conditional information, and outputs the denoised image.
Besides, it also uses AdaLN operation~\cite{Peebles2023DiT,yang2025moma} to inject time embedding into the denoising process.

\noindent\textbf{VLM as Instruction Encoder.}
WAN-2.1 originally uses umT5~\cite{chung2023unimaxfairereffectivelanguage} as its instruction encoder.
However, segmentation task needs to encode both images and text instructions, while umT5 is only capable of text encoding.
Thus, we replace the umT5 with an open-source vision-language model (VLM), Qwen2.5-VL-7B~\cite{bai2025qwen25vltechnicalreport}, to encode instructions for both image generation and segmentation tasks.
Specifically, for the segmentation task, the input instruction is formatted as follows:

\textit{``[Image]. Please segment the \{target\} in the image.''}

\noindent We extract the hidden states from its final layer to serve as the conditional input for the subsequent diffusion model. 

\noindent\textbf{VAE as Low-level Representation.}
VLMs primarily capture high-level semantic features, while segmentation tasks require low-level information such as texture and color connectivity for accurate pixel-level prediction.
Inspired from those image editing models~\cite{liu2025step1x-edit,labs2025flux1kontextflowmatching} which use VAE representation as a low-level shortcut,
we introduce an additional VAE feature of the input image into DiT for segmentation task specifically.
This latent representation is concatenates with randomly sampled noise to form the DiT's input.
We set the time embedding of the raw VAE representation to zero during AdaLN layer, indicating that it represents a completely clean (i.e., noise-free) image.

\begin{figure}
  \vspace{-0.5em}
  \centering
  \includegraphics[width=0.9\linewidth]{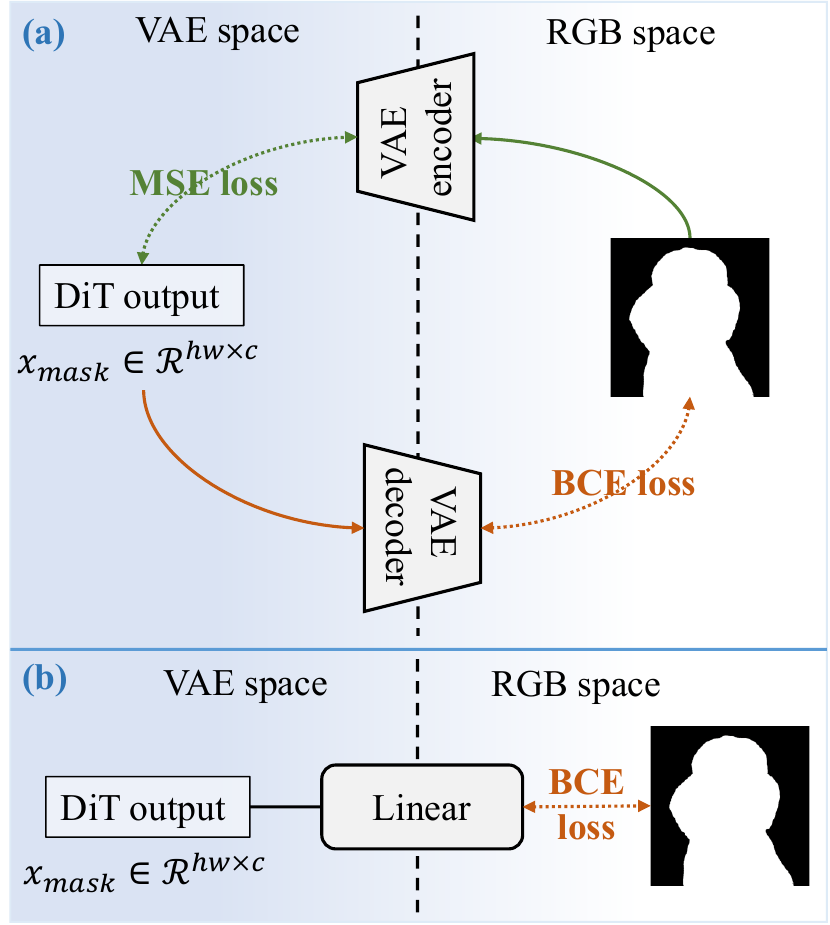}
  \caption{\textbf{Three variants of segmentation supervision formats.} \textbf{(a)} MSE loss in VAE space or BCE loss in RGB space with the help of VAE encoder-decoder. \textbf{(b)} By replacing the VAE decoder with a linear layer, we can apply BCE loss without backpropagation through frozen VAE.}
  \label{fig:supervize}
  \vspace{-1.5em}
\end{figure}

\noindent\textbf{Training Objective.}
Generative models commonly use mean squared error (MSE) in \cref{eq:mse-loss} for training, while binary segmentation tasks are typically optimized with binary cross-entropy (BCE) in label space. 
In this work we explore both supervision strategies and find MSE to be the preferred choice. It's simple to apply, incurs no extra decoder gradient flow, and tends to produce stronger results.

By contrast, applying BCE naively requires decoding VAE latents back to RGB and computing segmentation logits in pixel space, which forces gradients to flow through the VAE decoder.
This is inefficient and adds substantial computational overhead. 
As discussed in \cref{sec:seg_distribution}, VAE latents for segmentation masks are largely linearly separable. 
Motivated by this, we propose a third variant that replaces the VAE decoder with a simple learnable linear projection and applies BCE directly after this projection. 
This change removes the need to back-propagate through the full VAE decoder while preserving the ability to train with BCE. 
It also speeds up inference, since producing a mask only requires a single linear forward pass instead of the full decoder. \cref{fig:supervize} illustrates these three supervision pipelines; their comparative performance is analyzed in the ablation study \cref{sec:ablation-supervise}.

\noindent\textbf{CFG Process During Training \& Inference.}
Classifier-Free Guidance (CFG) is a conditioning technique that combines conditional and unconditional diffusion model scores to strengthen adherence to the conditioning signal without requiring an external classifier~\cite{ho2022classifierfreediffusionguidance}.
However, segmentation is inherently a deterministic prediction problem and therefore does not benefit from CFG.
As a result, we apply CFG only to natural-image generation during training, while segmentation samples remain strictly conditioned on the input image and textual instruction.
This design, in turn, allows segmentation masks to be produced with a single forward pass, eliminating the need for the dual conditional–unconditional evaluations required by CFG.

\section{Experiments}
\noindent\textbf{Implementation Details.}
Our framework is built upon the open sourced WAN-2.1 DiT model with 1.3B parameters~\cite{wan2025} and Qwen2.5-VL-7B VLM~\cite{bai2025qwen25vltechnicalreport}.
VLM and VAE encoder-decoders are kept frozen during training, while the whole DiT model is finetuned end-to-end with both segmentation and generation data.
We use cosine decay learning rate scheduler with initial learning rate 5e-5 and minimum learning rate 1e-5.
Most training settings converge around 8000 iterations with global batch size of 1024.
Segmentation and generation tasks are mixed with 1:1 ratio.

\noindent\textbf{Training Recipe.}
Our training recipe contains three types of data: semantic segmentation, referring segmentation, and text-to-image generation.
\textbf{(1) Semantic segmentation:} We use the COCO-stuff~\cite{coco_stuff}, ADE20K~\cite{ade20k} and PASCAL~\cite{pascal-voc-2010} dataset for semantic segmentation. These parts of data are reformatted into binary segmentation masks follow LISA~\cite{lisa}.
\textbf{(2) Referring segmentation:} We use the RefCOCO, RefCOCO+~\cite{referitgame} and RefCOCO-g~\cite{refcocog} dataset for referring segmentation.
\textbf{(3) Text-to-image generation:} We use open-sourced datasets such as DiffusionDB~\cite{diffusiondb}, BLIP-3o series~\cite{blip3o} as well as data provided from third party for text-to-image generation.

\noindent\textbf{Evaluation Metrics.}
Following conventions~\cite{lavt,vlt}, we evaluate our model on widely used referring segmentation benchmark RefCOCO series~\cite{lavt,vlt,mao2025safire}, using mIoU and oIoU metrics. We also report results on ReasonSeg~\cite{lisa}.

\subsection{Comparison with State-of-the-art Methods}

\noindent\textbf{Referring Segmentation Results.} 
Here we compare the performance of our approach with several state-of-the-art methods on referring segmentation benchmarks. The results are summarized in \cref{tab:res_benchmarks}, where we demonstrate the effectiveness and competitiveness of our model relative to existing approaches.

\begin{table}[htbp]
    \centering
    \caption{\textbf{Performance comparison on RES benchmarks.} The results are evaluated on RefCOCO, RefCOCO+~\cite{referitgame}, and RefCOCOg \cite{refcocog} (UMD partition) using mIoU and oIoU.}
    \label{tab:res_benchmarks}
    \vspace{-0.5em}
    \setlength{\tabcolsep}{3pt}
    \small
    \resizebox{\linewidth}{!}{%
    \begin{tabular}{c cccc}
        \toprule
        \multirow{2}{*}{\textbf{Method}} & \multicolumn{1}{c}{\textbf{RefCOCO}} & \multicolumn{1}{c}{\textbf{RefCOCO+}} & \multicolumn{1}{c}{\textbf{RefCOCO-g}} \\
        & \textbf{test A / B} & \textbf{test A / B} & \textbf{val / test} \\
        \midrule
        \multicolumn{4}{l}{\textit{Metric: oIoU}} \\ \midrule
        CRIS~\cite{cris}                  & 71.5 / 62.2 & 64.1 / 48.4 & 56.6 / 57.4 \\
        VLT~\cite{vlt}                    & 76.0 / 69.6 & 68.4 / 56.9 & 63.5 / 66.2 \\
        LAVT~\cite{lavt}                  & 75.8 / 68.8 & 68.4 / 55.1 & 61.2 / 62.1 \\
        BKINet~\cite{bkinet}              & 76.4 / 69.4 & 69.9 / 53.4 & 64.2 / 63.8 \\
        ReLA~\cite{grefcoco}              & 76.5 / 70.2 & 71.0 / 57.7 & 65.0 / 66.0 \\
        SLViT~\cite{slvit}                & 76.9 / 70.6 & 69.3 / 56.1 & 62.8 / 63.6 \\
        SADLR~\cite{sadlr}                & 76.3 / 70.1 & 69.1 / 55.2 & 63.6 / 63.6 \\
        DMMI~\cite{dmmi}                  & 77.1 / 70.2 & 69.7 / 57.0 & 63.5 / 64.2 \\
        CGFormer~\cite{cgformer}          & 77.3 / 70.6 & 71.0 / 57.1 & 64.7 / 64.1 \\
        RISCLIP~\cite{risclip}            & 76.5 / 69.8 & 70.6 / 55.5 & 64.1 / 65.1 \\
        MagNet~\cite{chng2024mask}        & 78.2 / 71.1 & 71.3 / 58.1 & 65.4 / 66.0 \\
        LQMFormer~\cite{lqmformer}        & 76.8 / 71.0 & 71.8 / 57.6 & 64.7 / 66.0 \\
        ReMamber~\cite{yang2025remamber}  & 76.7 / 70.9 & 70.8 / 57.5 & 63.9 / 64.0 \\
        MAGNET~\cite{chng2024mask}        & 78.3 / 72.2 & 73.6 / 61.8 & 67.8 / 69.3 \\
        PolyFormer-L~\cite{polyformer}    & 78.3 / 73.3 & 74.6 / 61.9 & 69.2 / 70.2 \\
        UNINEXT-L~\cite{uninext}          & 82.6 / 77.8 & 74.9 / 62.6 & 73.4 / 73.7 \\
        LISA~\cite{lisa}                  & 79.1 / 72.3 & 70.8 / 58.1 & 67.9 / 70.6 \\
        GLaMM~\cite{glamm}                & 83.2 / 76.9 & 78.7 / 64.6 & 74.2 / 74.9 \\
        u-LLaVA~\cite{ullava}             & 82.7 / 77.8 & 76.6 / 66.8 & 74.8 / 75.6 \\
        PSALM~\cite{zhang2025psalm}       & 78.1 / 76.6 & 70.7 / 64.4 & 71.0 / 72.3 \\
        GSVA~\cite{xia2024gsva}           & 80.4 / 74.2 & 71.5 / 60.9 & 74.2 / 75.6 \\
        PixelLM~\cite{pixellm}            & 76.5 / 68.2 & 71.7 / 58.3 & 69.3 / 70.5 \\
        \methodname (Ours)                 &\bf 83.3 / \bf 79.4 & \bf 78.7 / \bf 68.1 & \bf 75.6 / \bf 76.5 \\
        \midrule
        \multicolumn{4}{l}{\textit{Metric: mIoU}} \\ \midrule
        EEVG~\cite{eevg}                  & 79.6  / 75.3  & 75.6  / 64.6  & 71.5  / 71.9  \\
        PromptRIS~\cite{promptris}        & 81.2  / 74.6  & 76.6  / 64.3  & 69.2  / 70.5  \\
        OneRef-B~\cite{oneref}            & 81.9  / 77    & 77.9  / 69.6  & 74.1  / 74.9  \\
        C3VG~\cite{c3vg}                  & 82.9  / 79.1  & 79.6  / 72.4  & 76.3  / 77.1  \\
        \methodname (Ours)        &\bf 83.7 / \bf 80.7 & \bf 80.0 / \bf 73.1 & \bf 77.2 / \bf 78.2  \\
        \bottomrule
    \end{tabular}
    }
    \vspace{-1.25em}
\end{table}

\noindent\textbf{Reasoning Segmentation Results.}
Since our encoder is based on a vision-language model (VLM), it can also handle reasoning tasks as well. To fully leverage the VLM's reasoning capabilities during inference, we adopt a multi-stage pipeline. 
In the first stage, both the images and the instructions are provided to the VLM. The VLM outputs a clarified and more specific description of the target object to be segmented. 
In second stage, the refined instruction, together with the original image, is then passed to the DiT for segmentation inference. 
~\cref{tab:reasonseg_benchmarks} shows the performance of our model on ReasonSeg benchmarks.
\begin{table}[tbp]
    \centering
    \caption{\textbf{Performance comparison on ReasonSeg benchmarks.} Here $^*$ denotes the model finetuned on ReasonSeg training dataset.}
    \vspace{-0.5em}
    \label{tab:reasonseg_benchmarks}
    \setlength{\tabcolsep}{7pt}
    \begin{tabular}{c cc cc cc cc}
    \toprule
    \multirow{2}{*}{\textbf{Model}} & \multicolumn{2}{c}{\textbf{Val Set}} & \multicolumn{2}{c}{\textbf{Test Set}} \\
    \cmidrule(lr){2-3} \cmidrule(lr){4-5}
    & gIoU & cIoU & gIoU & cIoU \\
    \midrule
    SEEM~\cite{zou2023segment}                               & 25.5 & 21.2 & 24.3 & 18.7 \\
    Grounded SAM~\cite{ren2024grounded}                       & 26.0 & 14.5 & 21.3 & 16.4 \\
    OVSeg~\cite{liang2023open}                              & 28.5 & 18.6 & 26.1 & 20.8 \\
    GLaMM~\cite{rasheed2024glammpixelgroundinglarge}                      & 47.4 & 47.2 & --   & --   \\
    SAM4MLLM~\cite{chen2024sam4mllmenhancemultimodallarge}               & 46.7 & 48.1 & --   & --   \\
    LISA~\cite{lisa}                      & 44.4 & 46.0 & 36.8 & 34.1 \\
    LISA$^*$~\cite{lisa}                & \bf 52.9 & \bf 54.0 & 47.3 & 34.1 \\
    Ours               & 51.1 & 50.9&  \bf 52.3 & \bf 45.8 \\
    \bottomrule
    \end{tabular}
    \vspace{-2em}
\end{table}

\subsection{Visualization}
\begin{figure*}
    \centering
    \includegraphics[width=0.95\linewidth]{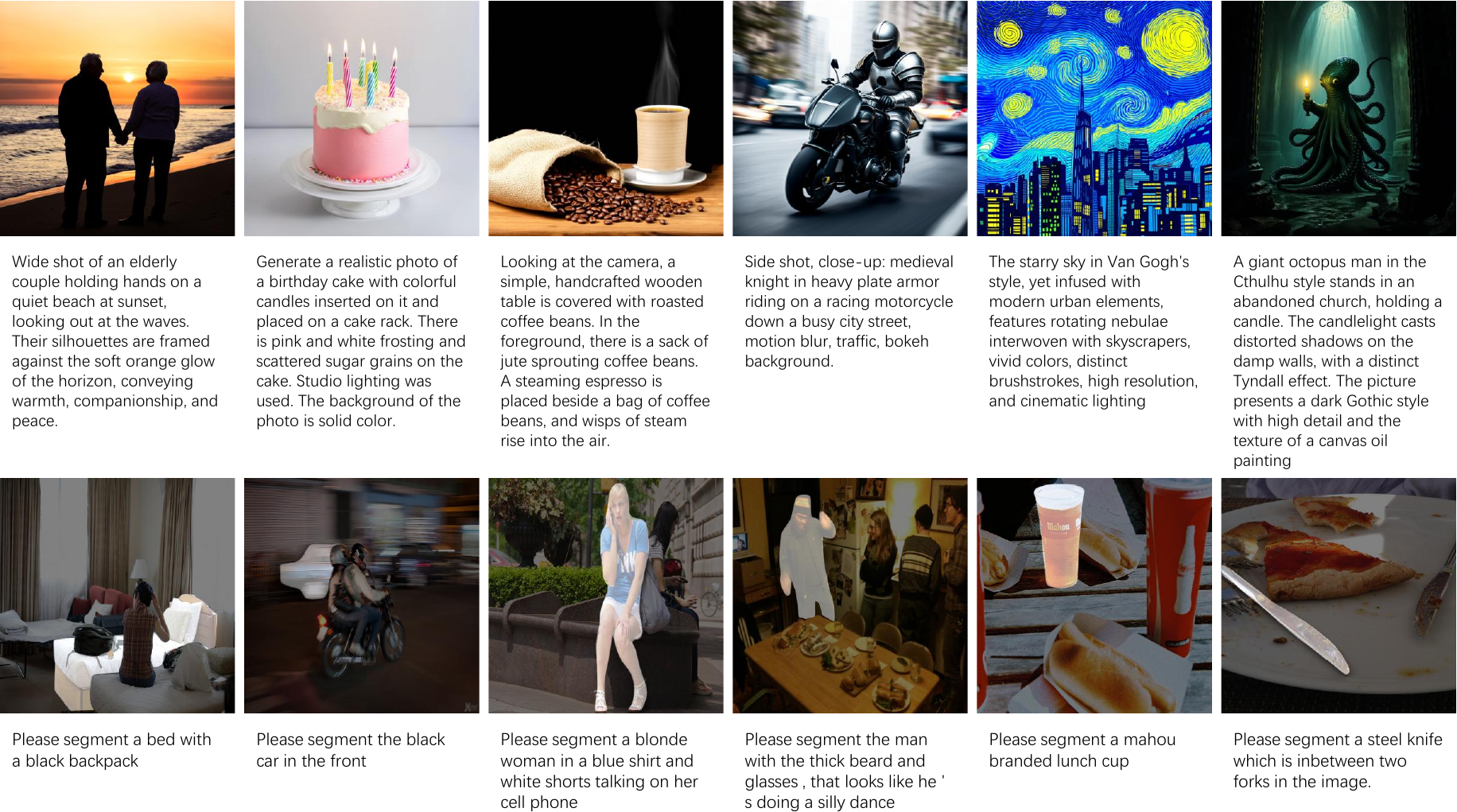}
    \caption{\textbf{Visualization results of our model.} We are able to generate both colorful images (up) and binary masks (down) simultaneously. For segmentation, we apply the output binary mask on the original image for better visualization.}
    \label{fig:visualization}
\end{figure*}

The visualization results of our model are presented in \cref{fig:visualization}, demonstrating its capability to simultaneously generate both colorful images and binary masks. For the segmentation outputs, the predicted binary masks are overlaid on the original images for clearer visualization.

\subsection{Ablation Studies}

\noindent\textbf{Sampling Strategy.}
We conduct an ablation study on the sampling strategy for the segmentation task by adjusting the hyperparameter $a$ in \cref{eq:seg-sample} which controls the degree of concentration towards the tail of the distribution. ~\cref{fig:sampling2_3} illustrates the sampling distributions for different values of $a$. 
Specifically, we experiment with $a\in[0.05, 0.1, 0.5]$.

\begin{table}[htbp]
  \centering
  \caption{\textbf{Performance comparison of different sampling strategies for segmentation task.}
    Adjusting segmentation sampling strategy to differ from generation is essential. A relatively smooth distribution (larger $a$) leads to degraded model performance,
    whereas the most extreme long-tailed distribution ($a=0.05$) achieves the best results.
  }
  \label{tab:ablation_sample}
  \setlength{\tabcolsep}{5pt}
  \begin{tabular}{c cc cc cc}
    \toprule
    \multirow{2}{*}{$a$ value} & \multicolumn{2}{c}{RefCOCO} & \multicolumn{2}{c}{RefCOCO+} & \multicolumn{2}{c}{RefCOCO-g} \\
    \cmidrule(lr){2-3} \cmidrule(lr){4-5} \cmidrule(lr){6-7}
    & mIoU & oIoU & mIoU & oIoU & mIoU & oIoU \\
    \midrule
    0.05 & 82.2 & 81.3 & 75.8 & 73.5 & 77.7 & 76.0 \\
    0.1  & 78.1 & 77.6 & 69.3 & 68.1 & 73.7 & 72.3 \\
    0.5  & 66.0 & 66.0 & 52.7 & 53.3 & 57.5 & 56.6 \\
    \bottomrule
  \end{tabular}
  \vspace{-1em}
\end{table}

The experimental results in~\cref{tab:ablation_sample} highlight that adapting the sampling strategy is essential for the effective training of the segmentation model.
While a larger $a$ ($a=0.5$) produces a smooth sampling distribution akin to generative training, it leads to the poorest performance. In contrast, a more long-tailed distribution ($a=0.1$) substantially improves results, with the most extreme strategy ($a=0.05$) achieving the best performance. This demonstrates that emphasizing high-noise samples is crucial for effective segmentation training.

\noindent\textbf{Segmentation Supervision Format.}
\label{sec:ablation-supervise}
As discussed in \cref{sec:architecture}, we propose three variants of segmentation supervision formats.
~\cref{fig:supervize} shows the pipeline of these three variants.
\cref{tab:ablation_loss} summarizes the performance. 
\textbf{(1)} We find that using MSE loss in the VAE space yields the best results, since it's the most closely aligned one with the original DiT training objective, 
thereby minimizing the need for additional adaptation. 
\textbf{(2)} Directly applying BCE loss in RGB space produces only moderate results. 
That's because, although the VAE encoder and decoder are kept frozen, applying BCE loss in the RGB space still requires backpropagation through the VAE, which makes optimization more difficult and reduces training efficiency. 
\textbf{(3)} Replacing the VAE decoder with a linear layer alleviates the optimization difficulty, but the performance still falls short of the MSE loss approach.

\vspace{-0.3em}
\begin{table}[htbp]
    \centering
    \caption{\textbf{Performance comparison of different segmentation supervision formats.} 
    MSE performs best since it's the most closely aligned with the original DiT training objective.
    Raw BCE performs poorly because it requires backpropagation through VAE, making optimization more difficult.
    Replacing the VAE decoder with a linear layer alleviates the optimization difficulty.
    }
    \vspace{-0.3em}
    \label{tab:ablation_loss}
    \setlength{\tabcolsep}{5pt}
    \begin{tabular}{c cc cc cc}
    \toprule
    \multirow{2}{*}{Loss} & \multicolumn{2}{c}{RefCOCO} & \multicolumn{2}{c}{RefCOCO+} & \multicolumn{2}{c}{RefCOCO-g} \\
      \cmidrule(lr){2-3} \cmidrule(lr){4-5} \cmidrule(lr){6-7}
    & mIoU & oIoU & mIoU & oIoU & mIoU & oIoU \\
    \midrule
    MSE &82.2& 81.3& 75.8& 73.5 & 77.7& 76.0 \\
    BCE &78.1& 77.6& 73.1& 70.3 & 75.0& 72.2 \\
    {\begin{tabular}[c]{@{}c@{}}BCE w/\\linear\end{tabular}} &81.3& 80.4& 74.8& 72.4 & 76.9& 75.0 \\
    \bottomrule
    \end{tabular}
    \vspace{-1em}
\end{table}

\noindent\textbf{Mix Training.}
Since our approach formulates the segmentation task within a generative training framework, data used for training generative models can be seamlessly integrated into the segmentation training stage. Consequently, we introduce text-to-image data at a 1:1 ratio for joint training. As shown in \cref{tab:ablation_mixtrain}, this incorporation of generative data leads to a positive improvement. The results indicate that such data augmentation is beneficial to segmentation performance, suggesting that the gap between generative modeling and segmentation may be less pronounced.

\begin{table}[htbp]
    \centering
    \caption{\textbf{Ablation study on mix training.} Adding generation data in an 1:1 ratio is beneficial to segmentation performance.}
    \label{tab:ablation_mixtrain}
    \setlength{\tabcolsep}{5pt}
    \begin{tabular}{c cc cc cc}
    \toprule
      \multirow{2}{*}{Gen data} & \multicolumn{2}{c}{RefCOCO} & \multicolumn{2}{c}{RefCOCO+} & \multicolumn{2}{c}{RefCOCO-g} \\
      \cmidrule(lr){2-3} \cmidrule(lr){4-5} \cmidrule(lr){6-7}
      & mIoU & oIoU & mIoU & oIoU & mIoU & oIoU \\
      \midrule
    $\checkmark$ & 82.2& 81.3& 75.8& 73.5 & 77.7& 76.0 \\
    $\times$ & 81.0& 80.6& 74.2& 72.4 & 76.7& 75.0 \\
    \bottomrule
    \end{tabular}
    \vspace{-1em}
\end{table}

\noindent\textbf{VAE Shortcut for Segmentation.}
Segmentation is actually a task that requires on low-level information, whereas the VLM component mainly captures semantic-level information. Therefore, we introduce a VAE-encoded image latent representation into the input of DiT to incorporate low-level features.
As shown in \cref{tab:ablation_vae}, the segmentation performance deteriorates significantly when the VAE input is removed. This marked decline underscores the critical role of low-level information provided by the VAE encoder.

\begin{table}[htbp]
    \centering
    \caption{\textbf{Ablation study on VAE input for segmentation task.} VAE input is crucial for segmentation task, it provides low-level information which is essential for accurate pixel-level prediction.}
    \label{tab:ablation_vae}
    \begin{tabular}{c cc cc cc}
    \toprule
      \multirow{2}{*}{VAE} & \multicolumn{2}{c}{RefCOCO} & \multicolumn{2}{c}{RefCOCO+} & \multicolumn{2}{c}{RefCOCO-g} \\
      \cmidrule(lr){2-3} \cmidrule(lr){4-5} \cmidrule(lr){6-7}
      & mIoU & oIoU & mIoU & oIoU & mIoU & oIoU \\
      \midrule
    $\checkmark$ & 82.2& 81.3& 75.8& 73.5 & 77.7& 76.0 \\
    $\times$ & 74.1& 73.2& 68.8& 67.6 & 72.4& 71.1 \\
    \bottomrule
    \end{tabular}
    \vspace{-1em}
\end{table}

\section{Related Work}
\noindent\textbf{Latent Diffusion Models and Their Representations.}
Latent Diffusion Models (LDM) \cite{ho2020denoising, rombach2022high, podell2023sdxl} perform diffusion in a compressed latent space, greatly improving generation quality and efficiency. Recent advances replace U-Net denoisers with transformers \cite{Peebles2023DiT, Chen2023PixArtAlpha, Esser2024SD3}, scaling successfully to text-to-image~\cite{betker2023dalle3, Chen2024PixArtSigma}, text-to-video \cite{Brooks2024Sora, DeepMind2025Veo3TechReport}, as well as unified generation and editing
frameworks~\cite{labs2025flux1kontextflowmatching,wang2026unireason,wu2025omnigen2,wang2026deepgen}.

Beyond generation, a growing body of work reveals that diffusion models also learn powerful representations \cite{Fuest2024DiffusionSurvey}: intermediate features exhibit strong discriminative ability \cite{Xiang2023DDAE, li2023your}, hybrid architectures jointly handle generation and recognition \cite{Yang2022HybridViTDiffusion, Tian2024ADDP}, and distillation pipelines transfer such representations to downstream tasks \cite{Yang2023RepLearner, Li2023DreamTeacher}.

\noindent\textbf{Generative Models for Segmentation.}
DatasetGAN~\cite{zhang2021datasetgan} and BigDatasetGAN~\cite{li2022bigdatasetgan} use GANs as labeled-data factories, producing unlimited image–mask pairs from a few annotations.
With the rise of diffusion models, many works extract their internal features for segmentation via frozen feature decoding~\cite{odice_diffusion,zhao2023vpd}, diffusion inversion~\cite{ma2023diffusionseg}, multi-step activation aggregation~\cite{luo2023dhf,meng2024diffusionmodelactivationsevaluated,lee2024dmp}, or feature distillation~\cite{fundel2025distilldift,stracke2025cleandift}.
These methods all treat diffusion models as implicit feature backbones with external decoders, leaving a gap with the generative pretraining objective. 

\vspace{-1ex}
\section{Conclusion}
We present \methodname, which argues for training segmentation in a generative style, by directly treating mask production as a conditional generation problem. 
Our investigation reveals that the VAE latents of binary masks differ from those of natural images,
and we bridge this gap by introducing a separate timestep sampling strategy that enables conflict-free joint training of segmentation and generation.
This formulation removes bespoke feature extraction pipelines, closes the optimization gap between pretraining and downstream adaptation, and naturally incorporates generated data to benefit segmentation.
Experiments on referring and reasoning segmentation benchmarks demonstrate \methodname's state-of-the-art performance.
Future work includes scaling the approach to larger DiT backbones and extending this unified generative paradigm to broader visual and multimodal tasks, such as medical image
segmentation~\cite{ma2024medsam,ma2021bsiris,liu2022tis} and audio-visual
segmentation~\cite{zhou2022avs,liu2023annotationfree,liu2023autr}.

\section*{Acknowledgments}
This work is supported by National Key R\&D Program of China (No. 2022ZD0160702), National Natural Science Foundation of China (No. 62306178), STCSM (No. 22DZ2229005), 111 plan (No. BP0719010).

{
    \small
    \bibliographystyle{ieeenat_fullname}
    \bibliography{main}
}


\end{document}